\documentclass[sigconf]{acmart}
\usepackage{amsmath} 
\usepackage{algorithm} 
\usepackage{algorithmic}
\usepackage{pgfplots}
\usepackage{multirow}
\usepackage{subcaption}

\AtBeginDocument{%
  \providecommand\BibTeX{{%
    \normalfont B\kern-0.5em{\scshape i\kern-0.25em b}\kern-0.8em\TeX}}}

\usepackage{titlesec}
  
\begin{document}

\renewcommand\footnotetextcopyrightpermission[1]{}

\title{PromptLA: Towards Integrity Verification of Black-box Text-to-Image Diffusion Models}

\author{Zhuomeng Zhang}
\email{zzmsmm@sjtu.edu.cn}
\affiliation{%
  \institution{Shanghai Jiao Tong University}
  \city{Shanghai}
  \country{China}
}

\author{Fangqi Li}
\email{solour_lfq@sjtu.edu.cn}
\affiliation{%
  \institution{Shanghai Jiao Tong University}
  \city{Shanghai}
  \country{China}
}

\author{Chong Di}
\email{cdi@qlu.edu.cn}
\affiliation{%
  \institution{Qilu University of Technology}
  \city{Jinan}
  \country{China}
}

\author{Hongyu Zhu}
\email{hongyu_z@sjtu.edu.cn}
\affiliation{%
  \institution{Shanghai Jiao Tong University}
  \city{Shanghai}
  \country{China}
}

\author{Hanyi Wang}
\email{why_820@sjtu.edu.cn}
\affiliation{%
  \institution{Shanghai Jiao Tong University}
  \city{Shanghai}
  \country{China}
}

\author{Shilin Wang}
\email{wsl@sjtu.edu.cn}
\affiliation{%
  \institution{Shanghai Jiao Tong University}
  \city{Shanghai}
  \country{China}
}

\begin{abstract}
Despite the impressive synthesis quality of text-to-image (T2I) diffusion models, their black-box deployment poses significant regulatory challenges: Malicious actors can fine-tune these models to generate illegal content, circumventing existing safeguards through parameter manipulation.
Therefore, it is essential to verify the integrity of T2I diffusion models. 
To this end, considering the randomness within the outputs of generative models and the high costs in interacting with them, we discern model tampering via the KL divergence between the distributions of the features of generated images. 
We propose a novel prompt selection algorithm based on learning automaton (PromptLA) for efficient and accurate verification.
Evaluations on four advanced T2I models (e.g., SDXL, FLUX.1)  demonstrate that our method achieves a mean AUC of over 0.96 in integrity detection, exceeding baselines by more than 0.2, showcasing strong effectiveness and generalization. Additionally, our approach achieves lower cost and is robust against image-level post-processing.
To the best of our knowledge, this paper is the first work addressing the integrity verification of T2I diffusion models, which establishes quantifiable standards for AI copyright litigation in practice.
\end{abstract}

\maketitle
\section{Introduction}
\label{sec:intro}

\begin{figure}[t]
    \centering
    \includegraphics[width=\linewidth]{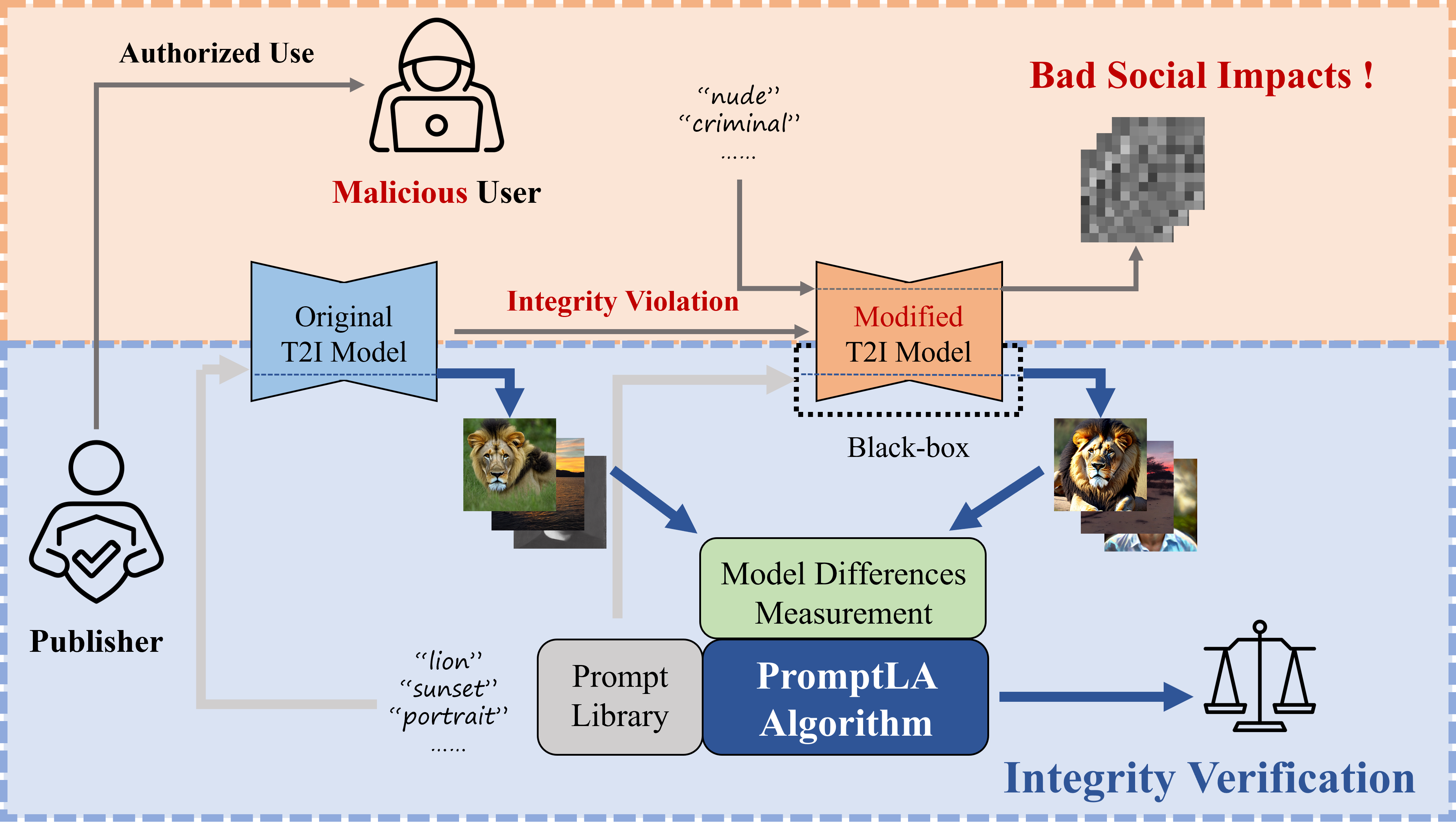}
    \caption{Integrity Violation Scenarios and proposed integrity verification framework of T2I diffusion models.}
    \label{fig:framework}
\end{figure}

Generative artificial intelligence has made significant progress in recent years. 
Notably, text-to-image (T2I) models~\cite{ramesh2021zero, nichol2021glide, ding2021cogview,  saharia2022photorealistic, ramesh2022hierarchical, li2024hunyuan}, exemplified by Stable Diffusion (SD)~\cite{rombach2022high}, have found widespread applications. 
To generate unique and stylized images, methods including Textual Inversion~\cite{gal2022image}, DreamBooth~\cite{ruiz2023dreambooth} and LoRA~\cite{ryu2023low} have been proposed as convenient tuning schemes for T2I diffusion models. 
However, malicious users can also take advantage of these techniques to breach the integrity of models even when they are not authorized to do so. For example, an attacker can inject celebrity faces, violent elements, or false information into a T2I model through LoRA training, thereby generating misleading images. Therefore, in the commercial scenario where the model is only authorized for using as it is, it is essential to verify the integrity of T2I diffusion models to ensure that users do not modify the model through any technique. Upon the occurrence of misconducts, an integrity verification method should attribute malicious behaviors to the user rather than the model's original owner, as shown in Figure~\ref{fig:framework}.

However, research on the integrity of T2I diffusion models is relatively lacking, with most work focusing on ownership verification, such as Stable Signature~\cite{fernandez2023stable} and Tree-ring~\cite{wen2023tree}. In the field of integrity verification, the focus has remained on relatively basic classification tasks. 

Compared with the integrity verification of models for classification tasks, integrity verification of T2I diffusion models has the following characteristics: (i) \textbf{randomness}, (ii) \textbf{complexity}, and (iii) \textbf{high access costs}, as shown in Table~\ref{tab:dnnvst2i}.
The integrity of classifiers can be examined by whether their predictions change for some trigger samples or not. 
However, due to the stochastic nature of the diffusion process, it is impossible to determine the integrity of models by simply comparing generated images for a fixed prompt. 
Moreover, generated images are inherently more complex compared to labels, and accessing the generated images of the model requires more time, which increases the difficulty of integrity verification. Reflecting modifications to models through generated images and selecting prompts for efficient integrity verification are key challenges in achieving integrity verification of T2I diffusion models.

To address the aforementioned challenges in verifying the integrity of T2I diffusion models, this paper makes the following contributions:

\begin{itemize}
    \item We introduce the first integrity verification framework for T2I diffusion models. We discern model modification via the KL divergence between the distributions of the features of generated images.
    \item We explore the impact of prompts on the feature distribution differences in generated images, and propose a prompt selection algorithm based on learning automaton to achieve precise and efficient integrity verification.
    \item Comprehensive experiments demonstrate that our algorithm achieves a mean detection AUC exceeding 0.95 across diverse integrity violations,  and remains robust against image-level post-processing.
\end{itemize}

\section{Related Work}
\subsection{Integrity Verification}

During the integrity verification, a malicious user obtains the original model $f_{0}$ and modifies it using some strategy $m$, resulting in a modified model $f_{m}$. The verifier can be the original model $f_{0}$ owner or a third-party institution, aiming to detect integrity violations by distinguishing between $f_{0}$ and $f_{m}$.

In the white-box scenario, integrity violations are detected by directly comparing the hash values of weights~\cite{zhu2022perceptual} of $f_{0}$ and $f_{m}$, after eliminating the structural symmetries~\cite{li2023linear}: 
\begin{equation}
    \texttt{hash}(f_{m}) \neq \texttt{hash}(f_{0}) \Rightarrow f_{m} \neq f_{0}.
\end{equation}

In the black-box scenario, the internal weights cannot be accessed so the hash of a model is intractable. 
Instead, two models are judged to be different only if their performance can be differentiated. 

In the field of integrity verification, the focus has remained on relatively basic classification tasks. 
Typical integrity violations of models for classification tasks include pruning~\cite{wu2021adversarial}, fine-tuning~\cite{cetinic2018fine}, etc. 
The integrity verification of classifiers uniformly relies on their outputs on a series of triggers $\mathbf{T}=\{\mathbf{t}_{n}\}^{N}_{n=1}$, which constitute their fragile fingerprints. 
If $f_{m}$ disagrees with $f_{0}$ on at least one trigger then the integrity violation is detected: 
\begin{equation}
    \exists \mathbf{t}_{n}\in \mathbf{T}, f_{m}(\mathbf{t}_{n}) \neq f_{0}(\mathbf{t}_{n}) \Rightarrow f_m \neq f_0.
\end{equation}
The triggers $\mathbf{T}$ should be samples sensitive to model changes~\cite{he2019sensitive, yin2023decision}, e.g., samples close to the decision boundary~\cite{aramoon2021aid, wang2023publiccheck}.

\subsection{Attribution of AI-generated images}
The detection of AI-generated images has long been a focal point in academic research. Various methods, including those based on color features~\cite{mccloskey2018detecting}, frequency features~\cite{durall2020watch}, gradient map extracted from pre-trained models~\cite{tan2023learning}, noise patterns derived from pre-trained denoising models~\cite{liu2022detecting} and features extracted from pre-trained CLIP:ViT model~\cite{ojha2023towards}, aim to identify and extract distinguishing characteristics of AI-generated images to differentiate them from real images.

Recently, researchers have proposed a more advanced objective—not only detecting AI-generated images but also attributing them to their generative models. DE-FAKE~\cite{sha2023fake} introduces two approaches: (1) utilizing ResNet-18~\cite{he2016deep} to classify the unique fingerprints left by different generative models on their output images, and (2) leveraging CLIP~\cite{radford2021learning} to extract features from images and their corresponding text, followed by training an MLP as the classifier. ManiFPT~\cite{song2024manifpt} derives artifacts for tracing by computing residuals based on the real data manifold. SemGIR~\cite{yu2024semgir} employs a semantic-guided image regeneration approach to decouple semantic information, thereby enabling attribution.

These attribution methods follow a feature extraction-classifier framework, can also be used for detecting integrity violations within the same model. However, these approaches require a substantial dataset to train the classifier. Moreover, they exhibit limited generalization, as in real-world scenarios, defenders cannot anticipate all possible integrity violation methods.


\subsection{Learning Automaton} 

As a fundamental component of non-associative reinforcement learning, wherein the environment operates independently of the input actions, Learning Automaton (LA) seeks to evaluate the efficacy of actions in an unknown environment through iterative interactions, ultimately identifying the optimal action among available choices~\citep{narendra2012learning}. 
Owing to its adaptive capabilities, LA has been extensively employed in various applications, including mathematical optimization~\citep{di2023learning}, pattern recognition~\citep{savargiv2022new}, cybersecurity~\citep{ayoughi2024enhancing}, and data mining~\citep{chakraborty2021learning}.

\begin{table}[t]
    \centering
    \fontsize{9}{10}\selectfont
\begin{tabular}{cccc}
\toprule
\textbf{Task}           & \begin{tabular}[c]{@{}c@{}}\textbf{Randomness}\\ \textbf{of outputs}\end{tabular} & \begin{tabular}[c]{@{}c@{}}\textbf{Complexity}\\ \textbf{of outputs}\end{tabular} & \begin{tabular}[c]{@{}c@{}}\textbf{Query}\\ \textbf{cost}\end{tabular} \\ 
\midrule
Classification &  Deterministic   &  Simple   &  Low                                                  \\
Text-to-Image  & Random & Complex &   High                                        \\ 
\bottomrule
\end{tabular}
    \caption{Characteristics of T2I diffusion models compared to classification tasks.}
    \label{tab:dnnvst2i}
\end{table}

\begin{figure*}[t]
    \centering
    \includegraphics[width=\textwidth]{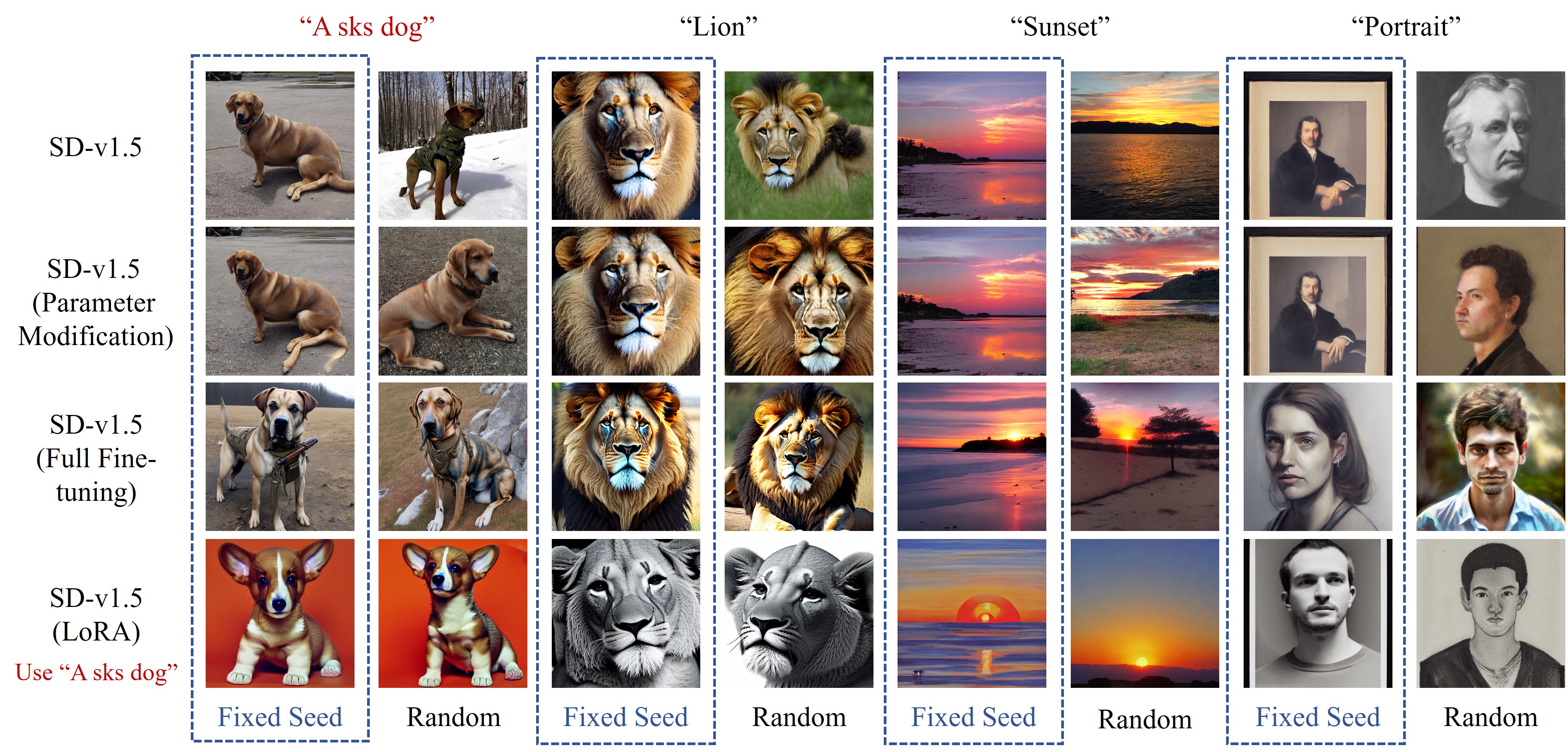}
    \caption{Comparison of images generated by the original model and after various integrity violations. For each prompt, the left column shows images generated with a fixed seed, and the right column shows randomly generated images.}
    \label{fig:Threat Model}
\end{figure*}

\section{Threat Model}
\noindent \textbf{Attackers' Assumptions}. Attackers, such as malicious users, aim to modify the original model $f_0$ to generate specific images, which can lead to bad effects. With white-box to $f_0$, attackers can compromise its integrity through methods such as full fine-tuning or parameter-efficient fine-tuning (PEFT), resulting in a modified model $f_m$. 

\noindent \textbf{Verifiers' Assumptions}. A Verifier, which could be either the model publisher or a third-party organization, aim to detect integrity violations. Following the approach of integrity verification in classification tasks, verifiers need to find a prompt $\textbf{p}$ such that $f_{m}($\textbf{p}$)$ can be distinguished from $f_{0}($\textbf{p}$)$. As shown in Figure~\ref{fig:Threat Model}, on the prompt which was used to conduct an integrity violation ("A sks dog" here), generated images from $f_m$ and $f_0$ can be easily distinguished. 
After eliminating the randomness by fixing the generator seed, a trivial comparison between images generated by different T2I diffusion models yields assertions on the integrity. 
However, the verifier cannot always know the prompts used to conduct integrity violations, and the same model might produce different images given the same prompt due to the randomness in diffusion process. In other words, the verifier has only black-box access to $f_m$.

Considering the characteristics of T2I diffusion models compared to classification tasks, as shown in Table~\ref{tab:dnnvst2i}, the integrity verification of T2I diffusion models should: 
(I) Measure the modifications to the underlying T2I diffusion models from the randomly generated images. 
(II) Cut down on the number of queries while ensuring the integrity violation can be detected with a high accuracy. (III) Be robust to various image post-processing operations and generalizable to different types of integrity violations.


\begin{figure}[htbp!]
\centering
\subfloat[v1.5-v1.5]{
		\includegraphics[width=0.489\linewidth]{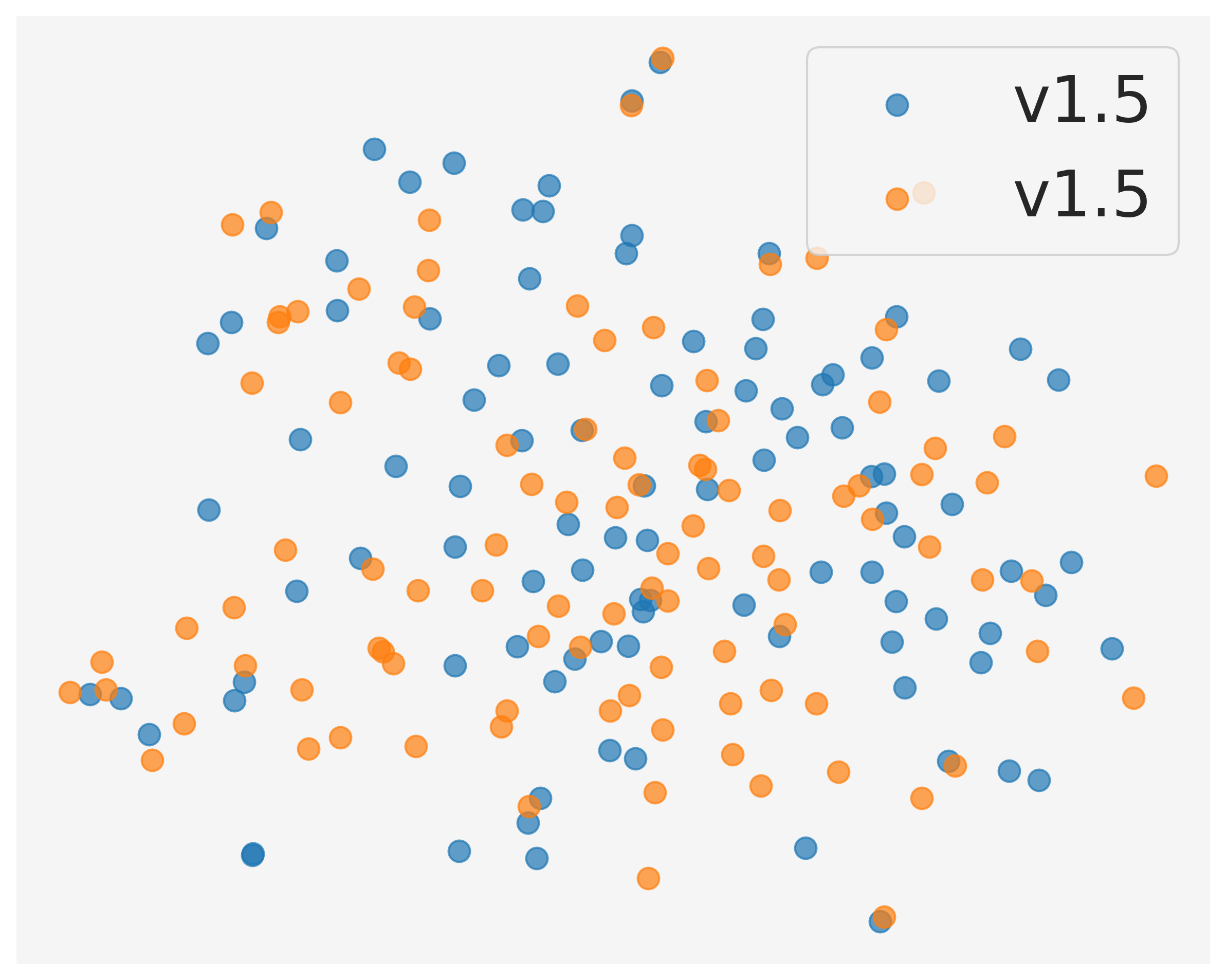}}
\hfill
\subfloat[v1.5-LoRA]{
		\includegraphics[width=0.489\linewidth]{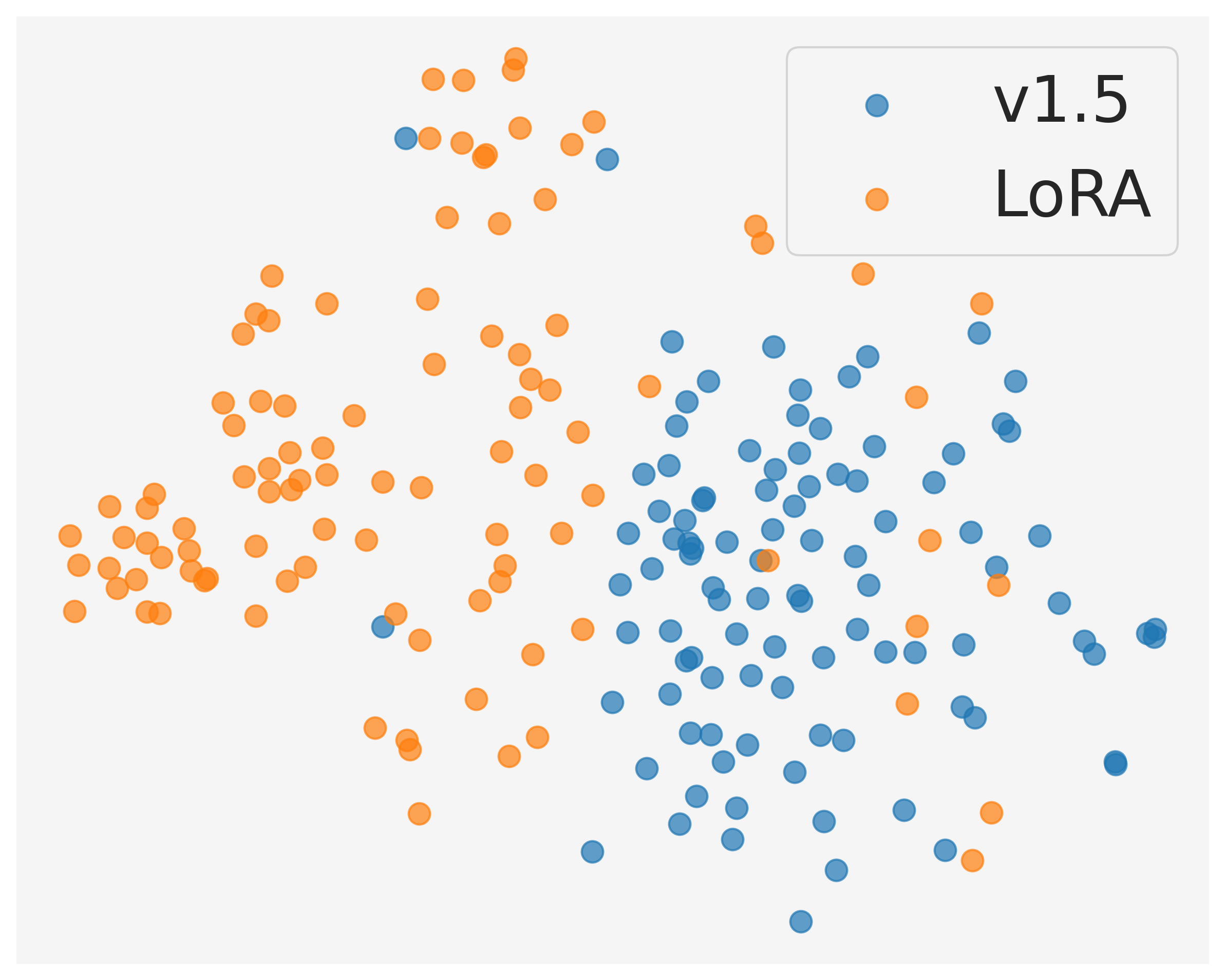}}
\caption{tSNE of features extracted from generated images using the Inception-v3 model and prompt7 "abstract". (a) Comparison within the original model SD-v1.5. (b) Comparison between the original model and the model fine-tuned using LoRA.}
\label{fig:tSNE}
\end{figure}

\begin{figure*}[t]
    \centering
    \includegraphics[width=1\textwidth]{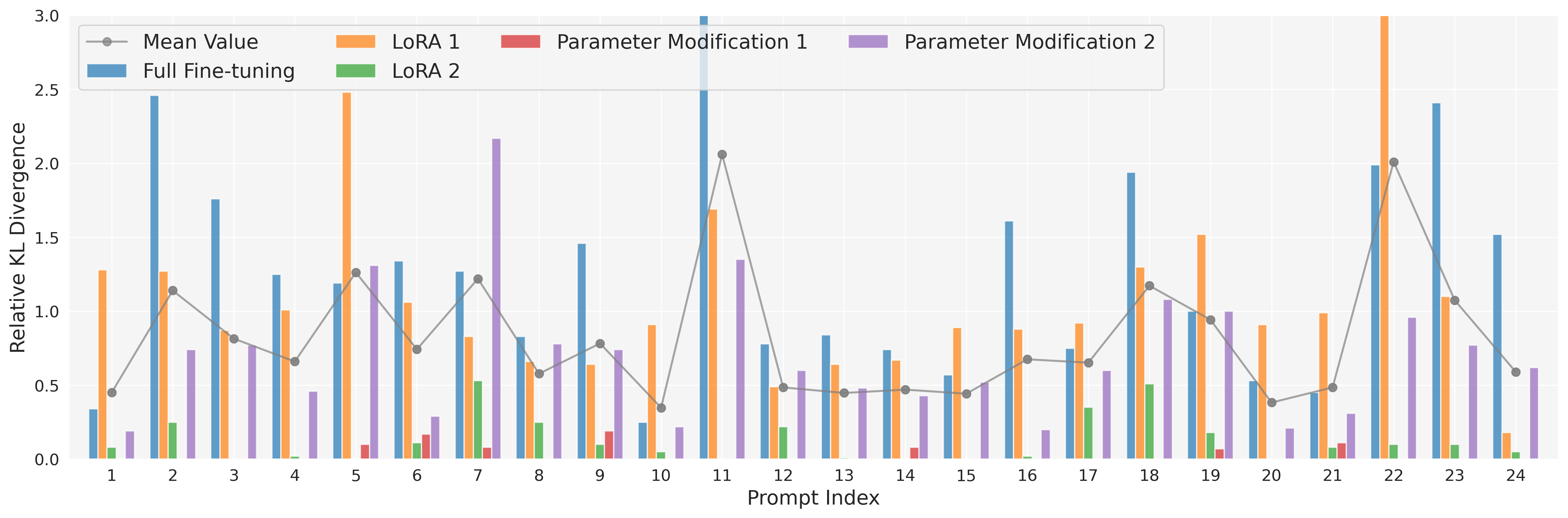}
    \caption{The relative KL divergence differences in the distribution of features extracted from images generated by T2I diffusion models before and after various integrity violations, using different prompt. The size of prompt library is set to 50 (24 of them are shown here), which is generated by GPT-4. The original model is SD-v1.5. Different colors represent different integrity violations, the details of which refer to the experiment section. The distribution of generated images' features is estimated using 50 images each. }
    \label{fig:RDD}
\end{figure*}

\section{Method}

\subsection{Model Differences Measurement}

To address the random nature of outputs from text-to-image (T2I) models, we follow the motivation that modifications to the model can be reflected in the feature distribution of generated images. 
This motivation is verified by an example shown in Figure~\ref{fig:tSNE}, where the feature distribution of images generated by models fine-tuned using LoRA differs from the feature distribution of images generated by the original model with the same prompt. 

To numerically measure the distance between randomly generated images from T2I diffusion models under the same prompt, we leverage the KL divergence ~\cite{hershey2007approximating} and employ a variational approach by assuming that the underlying distributions of features is a multivariate normal distribution~\cite{goodfellow2014generative, kingma2013auto, ho2020denoising} and further reduce the bias with standard
Bayesian estimation with non-informative prior~\cite{zyphur2015bayesian}. 

Formally, for the fixed prompt $\textbf{p}$ and T2I diffusion model $f$, we extract a collection of features $t_{\textbf{p}}(f)$ by querying $f$ for $n$ times:
\begin{equation}
\fontsize{9.5}{10}\selectfont
    t_{\textbf{p}}(f) = \{\mathcal{E}(\tau_l)|\tau_l \leftarrow f(\textbf{p})\}_{l=1}^n.
    \label{eq:extractor}
\end{equation}
where $\mathcal{E}$ is a feature extractor. Inception-v3~\cite{szegedy2016rethinking} pre-trained on ImageNet~\cite{deng2009imagenet} is used as $\mathcal{E}$. 
Given the original model $f_{0}$ and the suspicious model $f_{m}$, their outputs on a prompt $\textbf{p}$ are featured by $t_{\textbf{p}}(f_{0}), t_{\textbf{p}}(f_{m})$, which are then approximated by two multivariate normal distributions $P(t_{\textbf{p}}(f_0)) \sim \mathcal{N}(\mu_P, \Sigma_P)$ and $Q(t_{\textbf{p}}(f_m)) \sim \mathcal{N}(\mu_Q, \Sigma_Q)$ respectively. 
So the distance between two distributions can be computed by:
\begin{equation}
\fontsize{9.5}{10}\selectfont
    D_{KL}(P\|Q) = \int P(X) \log \frac{P(X)}{Q(X)} \, \text{d}X.
\label{eq:KL}
\end{equation}
To better compare different prompts, we use relative KL divergence to mitigate the impact of internal randomness on the KL divergence:
\begin{equation}
\fontsize{9.5}{10}\selectfont
    \beta_{\textbf{p}} = \frac{D_{KL}(P\|Q)}{D_{KL}(P\|P')} - 1.
    \label{eq:beta}
\end{equation}
where $P'$ is another independent estimation on the distribution of $t_{\textbf{p}}(f_{0})$. 

As shown in Figure~\ref{fig:RDD}, the distance between distributions of features of images generated by the model before and after integrity violations varies significantly with the prompt. 
Therefore, the integrity verification depends on maximizing the distance between distributions of features corresponding to certain prompts, which can be further reduced to selecting discriminating prompts. 

\begin{table}[h]
    \centering
    \fontsize{8}{9.5}\selectfont
    \begin{tabular}{c|l}
    \toprule
     \textbf{Symbol}    & \textbf{Explanation}  \\ 
     \midrule
     $\alpha$ & Significance level. \\
     $r$ & The number of round, $r\in\left\{1,2,\cdots,R\right\}$. \\
     $R_s$ & The number of round to start the hypothesis test. \\
     $R_e$ & The total number of rounds. \\
     $A(r)$ & Action set at the $r$-th round.\\
     $q$ & Size of $A$, $|A(1)| = |A| = q$. \\
     $\beta^r_i(k)$ & Feedback for $k$-th choose for action $a_i$ at the $r$-th round. \\ 
     $F_i(r)$   & Feedback sequence for action $a_i$ at the $r$-th round. \\
     $n$ & The number of images generated per prompt per round.\\
     $\hat{d}_i(r)$ & Mean of the feedback sequence $F_i(r)$.\\
     $a_{m(r)}$ & Estimated optimal action after the $r$-th round.\\
     \bottomrule
    \end{tabular}
    \caption{Notations used in PromptLA algorithm.}
    \label{tab:notations}
\end{table}

\subsection{PromptLA}
We remark that selecting the most discriminating prompt from a pool of candidates by interacting with black-box T2I diffusion models with the least number of queries is essentially a stochastic optimization task, which can be efficiently solved by reinforcement learning algorithms. 
We combine the learning automaton (LA) framework based on statistical hypothesis testing approach~\cite{di2019novel} and design a prompt selection algorithm (PromptLA). In a nutshell, an LA interacts with a stochastic environment (i.e., the feedback from the environment might be different even when the LA chooses the same action) by continually selecting actions, updating its strategy, and converging to the optimal action. 
In our setting, the environment consists of two T2I diffusion models $\left\{f_{0}, f_{m}\right\}$, while the set of actions is the set of candidate prompts. 
The notations that are used in defining the algorithm are summarized in Table~\ref{tab:notations}. 

\subsubsection{Action Set Construction}
We construct the pool of prompts $L=\{\textbf{p}_i\}^N_{i=1}$ with GPT-4~\cite{achiam2023gpt}. 
At the beginning, $q$ prompts that have not been examined before are randomly chosen from $L$ to form the action set $A$. 
The algorithm runs for multiple rounds and the action set at the $r$-th round is denoted by $A(r)$. 
Naturally, $|A(1)| = |A| = q$ and we set $q$ to 5.

\subsubsection{Feedback Calculation} 
At the $r$-th round, all the remaining actions in the action set $A(r)$ are chosen to interact with environment. 
To generate more feedback (thus reducing the bias in estimation the optimal prompt) with fewer queries, we conduct a cross-validation among all historical data. 
Concretely, for each action $a_i \in A(r)$, the feedback sequence $F_{i}(r)$ is appended with $r$ extra feedback $F_i(r) = F_i(r-1) \cup \left\{\beta^r_i(k)\right\}_{k=1}^{r}$, where $\beta^r_i(k)$ is the relative KL divergence computed by Eq.~\eqref{eq:beta}, in which $P,P',Q$ are estimated from $nk$ images produced by $f_{0},f_{0},f_{m}$ so far.

\subsubsection{Action Set Updating Strategy}
The action set updating strategy follows the statistical hypothesis testing proposed in ~\citet{di2019novel}. 
When the $r$-th round terminates, we compute the estimated reward probability $\hat{d}_i(r)$
\begin{equation}
\fontsize{9.5}{10}\selectfont
    \hat{d}_i(r) = \frac{\sum\limits^r_{l=1} {\sum\limits^l_{k=1} \beta^l_i(k)}}{|F_i(r)|}.
    \label{d}
\end{equation}

The estimated optimal action $a_{m(r)}$ after the $r$-th round is
the one in $A(r)$ with the highest estimated reward probability $\hat{d}_i(r)$, i.e.:
\begin{equation}
\fontsize{9.5}{10}\selectfont
    a_{m(r)} = \arg\max\limits_{a_i \in A(r)} \hat{d}_i(r).
    \label{a}
\end{equation}

\begin{algorithm}[t]
\caption{PromptLA}
\label{alg:algorithm}
\textbf{Input}: $A$, $f_0$, $f_m$.\\
\textbf{Parameter}: $\alpha$, $R_s$, $R_e$, $n$.\\
\textbf{Output}: $a_m$, $\hat{d}_m$.
\begin{algorithmic}[1] 
\STATE Let $r=1$, $A(1)=A$, $\forall a_i \in A, F_i(0) = \emptyset$.
\WHILE{$r\leq R_e$ and $|A(r)|>1$}
\FOR{$a_i \in A(r)$}
\STATE $F_i(r) = F_i(r-1)$
\FOR{$k = 1,2,\cdots,r$}
\STATE Compute feedback $\beta^r_i(k)$ by Eq.~\eqref{eq:beta}.
\STATE $F_i(r) = F_i(r) \cup \{\beta^r_i(k)\}$ 
\ENDFOR
\STATE Compute $\hat{d}_i(r)$ by Eq.\eqref{d}.
\ENDFOR
\STATE Find out estimated optimal action $a_{m(r)}$ by Eq.~\eqref{a}.
\IF {$r \geq R_s$}
\FOR{$a_i \in A(r), i \neq m(r)$}
\IF{$|F_i(r)| \leq 30$}
\STATE Given $F_{m(r)}(r)$, $F_i(r)$ and $\alpha$, implement the t-test.
\ELSE
\STATE Given $F_{m(r)}(r)$, $F_i(r)$ and $\alpha$, implement the Z-test.
\ENDIF
\IF{the null hypothesis $H_0: d_i = d_{m(r)}$ is rejected}
\STATE $A(r+1) = A(r) \setminus \{a_i\}$
\ENDIF
\ENDFOR
\ENDIF
\STATE $r = r + 1$
\ENDWHILE
\STATE \textbf{return} $a_m(r)$, $\hat{d}_m(r)$
\end{algorithmic}
\end{algorithm}

After $R_s$ rounds, the statistical hypothesis testing start. 
Following ~\citet{di2019novel}, the Student’s t-test is adopted when $|F_i(r)|\leq 30$, while the Z-test is adopted in later rounds. 
For each non-optimal action $a_i \in A(r), i \neq m(r)$, the statistical test is conducted given the feedback sequences of action $a_i$ and $a_{m(r)}$ and the significance level $\alpha$. 
If the null hypothesis $H_0: d_i = d_{m(r)}$ is rejected, action $a_i$ is
eliminated from the current action set. 
All the remaining actions in the action set
are explored in later rounds

We remark that one difference between integrity verification and ordinary optimization is that it is sufficient to find one prompt that can distinguish two models (rather than find the most discriminating one). 
There is no need to run the algorithm until convergence as in the traditional LA.
The entire process terminates after $\min(R, R_e)$ rounds and returns the most discriminating prompt at that time. 
The choice of $R_e$ reflects a trade-off between algorithm performance and time consumption.

The overall process of the proposed PromptLA is summarized in Algorithm~\ref{alg:algorithm}.

\subsection{Integrity Verification Framework}

As shown in Figure~\ref{fig:framework}, T2I diffusion models publisher can conduct the integrity verification when he/she suspects that a malicious user who has been authorized to use $f_0$ only for benign purposes modified the model for harmful purposes. 

For a suspicious model $f_m$, the publisher first constructs a prompt library $L=\{\textbf{p}_i\}^N_{i=1}$ and sets a threshold $\theta$. 
Then PromptLA tests the prompt library gradually. 
Each time, the algorithm selects $q$ prompts and returns the most discriminating prompt which is $a_m$ from action set, and its corresponding relative KL divergence $\hat{d}_m$. 
If $\hat{d}_m \geq \theta$ then an integrity violation is reported. Otherwise, PromptLA continues to explore the remaining prompts in the library. 
If, after traversing the prompt library, no $\hat{d}_m$ of prompts exceeds the set threshold $\theta$, the model's integrity is considered intact, and the highest value encountered during the process is recorded for the AUC calculation.

\begin{table*}[t]
    \centering
    \fontsize{7.5}{7.5}\selectfont
\begin{tabular}{ccccccccccccccc}
\toprule
\multirow{2}{*}{\textbf{Method}} &                    \multicolumn{6}{c}{\textbf{SD-v1.5}}                   & \multicolumn{3}{c}{\textbf{SDXL}} & \multicolumn{2}{c}{\textbf{HunyuanDiT-v1.2}} & \multicolumn{2}{c}{\textbf{FLUX.1-dev-fp8}}& \multirow{2}{*}{\textbf{Avg}} \\ \cmidrule(lr){2-7}\cmidrule(lr){8-10}\cmidrule(lr){11-12}\cmidrule(lr){13-14}
                        & FFT   & LoRA1\textsuperscript{1} & LoRA2 & PM1\textsuperscript{2}   & PM2   & v1.4  & LoRA1\textsuperscript{1}  & LoRA2  & PM\textsuperscript{2}     &v1.1\textsuperscript{1}                & PM\textsuperscript{2}             & LoRA1\textsuperscript{1} &LoRA2\textsuperscript{2}                     \\ 
                        \midrule
McClo18~\cite{mccloskey2018detecting}\textsuperscript{1}                 & 0.028 & \textbf{1.000} & 0.286 & 0.284 & 0.561 & 0.161 & \textbf{1.000}  & 0.316  & 0.451  & 1.000            & 0.003            &\textbf{1.000} &0.971 & 0.543                \\
Durall20~\cite{durall2020watch}\textsuperscript{1}                
& 0.035 & \textbf{1.000} & 0.218 & 0.253 & 0.074 & 0.363 & \textbf{1.000}  & 0.000  & 0.000  & \textbf{1.000}            & 0.543            & \textbf{1.000} &0.998 &0.499                \\
DE-FAKE~\cite{sha2023fake}\textsuperscript{1}                 
& 0.775 & \textbf{1.000} & 0.500 & 0.500 & 0.725 & 0.500 & \textbf{1.000}  & 0.500  & 0.500  & \textbf{1.000}            & 0.395            & \textbf{1.000} &0.505 &0.684                \\
ManiFPT~\cite{song2024manifpt}\textsuperscript{1}                 
& 0.325 & \textbf{1.000} & 0.350 & 0.425 & 0.838 & 0.425 & \textbf{1.000}  & 0.500  & 0.500  & \textbf{1.000}            & 0.160            & \textbf{1.000} &0.725 &0.634                \\
SemGIR~\cite{yu2024semgir}\textsuperscript{1}                  
& 0.525 & \textbf{1.000} & 0.500 & 0.500 & 0.525 & 0.500 & \textbf{1.000}  & 0.500  & 0.500  & \textbf{1.000}            & 0.381            & \textbf{1.000} &0.500 &0.649                \\ 
\midrule
McClo18~\cite{mccloskey2018detecting}\textsuperscript{2}                 & \underline{0.500} & 0.500 & 0.500 & 0.500 & 0.500 & 0.500 & 0.000  & 0.000  & \textbf{1.000}  & 0.000            & 0.997            & 0.998 & \textbf{1.000} & 0.538               \\
Durall20~\cite{durall2020watch}\textsuperscript{2}                
& \textbf{1.000} & 0.000 & 0.671 & 0.695 & 0.985 & 0.003 & \textbf{1.000}  & \textbf{1.000}  & \textbf{1.000}  & 0.028            & 0.998            & \textbf{1.000}  & \textbf{1.000} & 0.722              \\
DE-FAKE~\cite{sha2023fake}\textsuperscript{2}                 
& 0.377 & 0.000 & 0.806 & 0.379 & 0.589 & 0.436 & 0.225  & 0.350  & \textbf{1.000}  & 0.000            & \textbf{1.000}            & \textbf{1.000}  & \textbf{1.000} & 0.551              \\
ManiFPT~\cite{song2024manifpt}\textsuperscript{2}                 
& \textbf{1.000} & 0.000 & 0.368 & 0.574 & 0.449 & 0.348 & 0.300  & 0.100  & \textbf{1.000}  & 0.000            & \textbf{1.000}            & 0.500    & \textbf{1.000} & 0.511            \\
SemGIR~\cite{yu2024semgir}\textsuperscript{2}                  
& 0.648 & 0.000 & 0.971 & 0.643 & 0.705 & 0.415 & 0.275  & 0.300  & \textbf{1.000}  & 0.000            & \textbf{1.000}            & \textbf{1.000} & \textbf{1.000} & 0.612               \\ 
\midrule
PromptLA\_v1(Ours)      & \textbf{1.000} & 0.995 & 0.945 & \textbf{0.732} & 0.993 & 0.928 & \textbf{1.000}  & \textbf{1.000}  & \textbf{1.000}  & \textbf{1.000}            & \textbf{1.000}    &\textbf{1.000} &\textbf{1.000}        & 0.969                \\
PromptLA\_v2(Ours)      & \textbf{1.000} & \textbf{1.000} & \textbf{0.995} & 0.690 & \textbf{0.997} & \textbf{0.953} & \textbf{1.000}  & \textbf{1.000}  & \textbf{1.000}  & \textbf{1.000}            & \textbf{1.000}    &\textbf{1.000} &\textbf{1.000}        & \textbf{0.972}                \\ 
\bottomrule
\end{tabular}
    \caption{The verification AUC comparison between PromptLA and baselines. (1) Baseline methods\textsuperscript{1} were trained on datasets composed of negative samples generated from integrity-violated\textsuperscript{1} model. (2) Baseline methods\textsuperscript{2} were trained on datasets composed of negative samples generated from integrity-violated\textsuperscript{2} model. Each Baseline method independently trained a classifier for each base model and tested. The prompt chosen for baselines was prompt22 "Lion", which is more significant for most violations, as shown in Figure~\ref{fig:RDD}. The best results for each violation are denoted in boldface. The 'Avg' column represents the average AUC across all integrity violations for each method.}
    \label{tab:new_auc}
\end{table*}

\section{Experiments and Discussions}

\subsection{Settings}

\subsubsection{Base Models} The original T2I diffusion models to be modified should be fully open-source and widely used. We chose Stable Diffusion (SD)~\cite{rombach2022high}-v1.5, SDXL~\cite{podell2023sdxl}, HunyuanDiT-v1.2~\cite{li2024hunyuan} and FLUX.1-dev-fp8 as original models whose integrity need to be protected.

\subsubsection{Integrity Violations}

From the attacker's perspective, the most common model modification methods can be categorized into Full Fine-tuning and parameter-efficient fine-tuning (PEFT). We choose LoRA a representative of PEFT methods in this paper.
Other options include vanilla parameter modification or version rollback. 
They can hardly fulfill malicious purposes, but they serve as good examples of integrity violations for evaluation.
In summary, the integrity violations studied in this paper are:

\begin{itemize}
    \item \textbf{Full Fine-tuning (FFT)}: For SD-v1.5, we selected the open-source stylized model Dreamlike-diffusion-1.0 available on Hugging Face~\footnote{https://huggingface.co/dreamlike-art/dreamlike-diffusion-1.0}.
    \item \textbf{LoRA}: For SD-v1.5, firstly, we employed the samples and methodology of Dreambooth~\footnote{https://huggingface.co/docs/peft/main/en/task\_guides/dreambooth\_lora} for local fine-tuning, and secondly, we selected LoRA models published by others on Civitai~\footnote{https://civitai.com/}, which was also applied to SDXL and FLUX.1-dev-fp8. Different LoRA instances are distinguished by numerical suffixes.
    \item \textbf{Parameter Modification (PM)}: For SD-v1.5 and SDXL, we added various levels of random noise to the parameters of Unet, and the parameters of transformer for HunyuanDiT-v1.2. The noise is uniformly sampled from the [0, 1) interval and scaled by a coefficient that modulates its amplitude. Different PM instances are distinguished by numerical suffixes. 
    \item \textbf{Version Rollback}: For SD-v1.5, we selected the older version SD-v1.4, and for HunyuanDiT-v1.2, we selected the older version HunyuanDiT-v1.1.
\end{itemize}

\subsubsection{Baselines} 
For comparison, we transformed schemes designed for attribution of AI-generated images, such as \textbf{McClo18}~\cite{mccloskey2018detecting}, \textbf{Durall20}~\cite{durall2020watch}, \textbf{DE-FAKE}~\cite{sha2023fake}, \textbf{ManiFPT}~\cite{song2024manifpt} and \textbf{SemGIR}~\cite{yu2024semgir} into baselines for integrity verification. 
 Baselines for integrity verification use different features of generated images as training samples to train a binary classifier under known attacks, which is then tested on a test set. Following their settings, We used CNN for McClo18, MLP for Durall20, DE-FAKE and SemGIR, and ResNet50 for ManiFPT as the classifier after feature extraction for training. To make better comparisons, we trained the model using different training set whose negative samples were generated from different integrity violation as shown in Table~\ref{tab:new_auc}.

\subsubsection{Evaluation Metrics} 
Integrity verification schemes are evaluated in their AUC in the binary classification between intact T2I diffusion models from modified ones. 
The cost of a scheme is measured in the number of images generated by the T2I diffusion models until the verification process terminates. 

\subsubsection{Implementation Details} 
In PromptLA, the number of images generated per prompt per model per round was set to $n=5$, the starting round for filtering was $R_s=5$, the total number of rounds was $R_e=10$. 
We considered two configurations to the significance level $\alpha$, and the threshold $\theta$, as (0.01, 0.25) and (0.05, 0.3), which have good performance in comprehensive testing. 
These versions are denoted as \textbf{PromptLA\_v1} and \textbf{PromptLA\_v2}.

Two RTX 3080 GPUs were used for experiments and image generation. Using default parameters, it takes about 5 seconds to access the SD-v1.5 and generate an image, 14 seconds for SD-v2.1, 45 seconds for HunyuanDiT-v1.2 and 60 seconds for Flux.1-dev-fp8.

The training sets for baselines consist of 400 positive and 400 negative samples. All evaluations were repeated for 20 times for each different integrity violations to compute the AUC and the average cost. 

\subsection{Accuracy Evaluation}

Table~\ref{tab:new_auc} shows that our method outperformed the baselines across various integrity violations, especially in detecting fine-grained modifications that are hard to spot such as LoRA2, PM1 and PM2 for SD-v1.5. 

\begin{figure}[t]
    \centering
    \includegraphics[width=\linewidth]{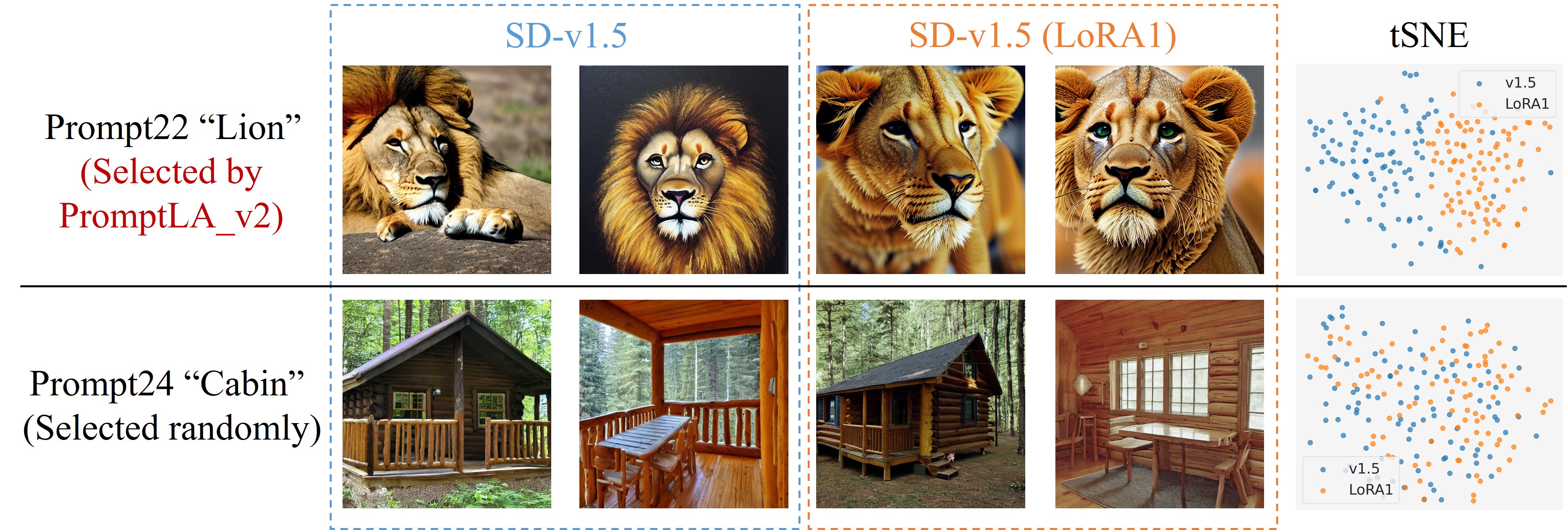}
    \caption{A visual instance of prompt selection using PromptLA against integrity violation LoRA1. Base model: SD-v1.5.}
    \label{fig:show}
\end{figure}

Baselines achieved good detection results for known attacks (i.e., the defender has known the modification that the adversary has performed), such as baselines trained on LoRA1 for SD-v1.5, PM for SDXL, v1.1 for HunyuanDiT-v1.2 and LoRA2 for FLUX.1-dev-fp8. However, they performed poorly on unknown attacks, which are not in their training set, indicating a lack of generalization. This aligns with our previous analysis: classifiers perform well in detecting integrity violations within the distribution of the training data but lack the capability to generalize across different types of integrity violations. For instance, methods such as DE-FAKE~\cite{sha2023fake}, ManiFPT~\cite{song2024manifpt} and SemGIR~\cite{yu2024semgir}, which employ more advanced feature extraction techniques, are actually more prone to overfitting, leading to significantly reduced AUC scores—sometimes even approaching zero—on other distributions.
 
In contrast, PromptLA selects the appropriate prompt for any integrity violation efficiently, which demonstrated strong performance and generalization capability. As shown in Figure~\ref{fig:show}, the prompt selected by PromptLA exhibited higher discriminative qualities than random prompts. 

If we remove the well-performing prompt22 "Lion" from the prompt library, which is shown in Figure~\ref{fig:RDD}, the performance of the baselines might be even worse. 
It is remarkable that they could still serve as promising integrity verification schemes against specific integrity violation, even though they were not designed for this purpose. 
However, due to the additional training costs involved, it is still less effective compared to the PromptLA algorithm.

We also find that more advanced and complex T2I diffusion models, such as SDXL, HunyuanDiT and FLUX, are easier to detect when their integrity are violated, as shown in Table~\ref{tab:new_auc}. We believe that the reason is that models with more complex architectures and a larger number of parameters embed more inherent characteristics in their generated images, which is also consistent with analyses in SemGIR~\cite{yu2024semgir}. Additionally, higher-quality generated images are more likely to reflect differences before and after integrity violations.

\begin{figure}[t]
\centering
\subfloat[PromptLA\_v1 (jpeg)]{
		\includegraphics[width=0.489\linewidth]{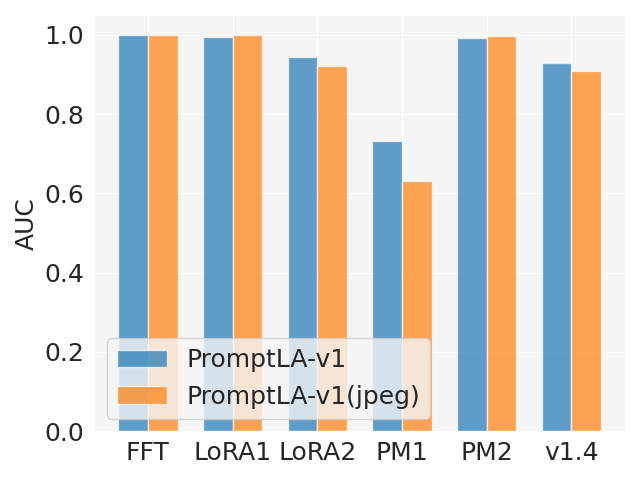}}
\hfill
\subfloat[PromptLA\_v2 (jpeg)]{
		\includegraphics[width=0.489\linewidth]{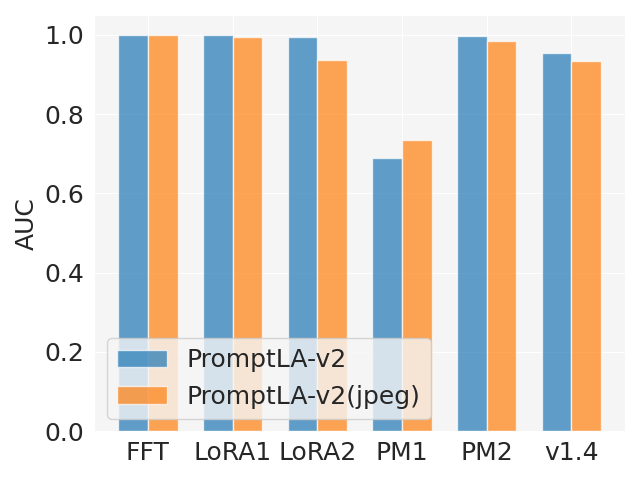}}
\\
\subfloat[PromptLA\_v1 (crop)]{
		\includegraphics[width=0.489\linewidth]{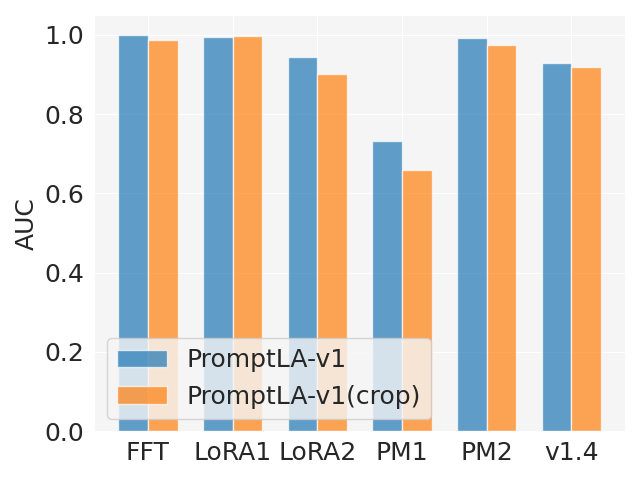}}
\hfill
\subfloat[PromptLA\_v2 (crop)]{
		\includegraphics[width=0.489\linewidth]{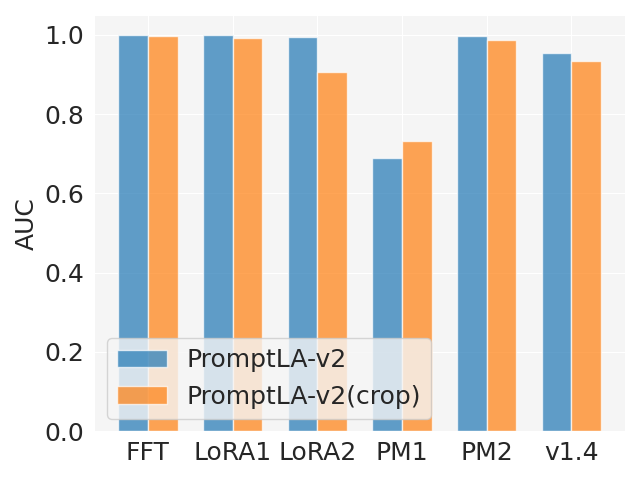}}
\caption{Robustness against image-level post-processing (cropping and jpeg compression). Base model: SD-v1.5.}
\label{fig:stable}
\end{figure}

\subsection{Robustness Evaluation}

In practice, image-level post-processing operations such as cropping, compression, and scaling might change the characteristics of images, leading to false integrity alarms. 
Therefore, an integrity verification algorithm should overlook such image-level post-processing. 
As shown in Figure~\ref{fig:stable}, when being confronted by random cropping and 85\% JPEG quality compression, both \textbf{PromptLA\_v1} and \textbf{PromptLA\_v2} maintained the high AUCs, indicating robust detection performance for various violations. 
Rotation and scaling were included in preprocessing of generated images, which had no significant impact on the results.

This demonstrates that the PromptLA algorithm's performance is stable under certain image-level post-processing such as random cropping and jpeg compression, thanks in part to the robustness of the image feature extraction model Inception-v3~\cite{szegedy2016rethinking}.

\subsection{Ablation Study}

We also evaluated the necessity of the PromptLA algorithm for integrity verification of T2I diffusion models through ablation studies. 
Without PromptLA, we employed a vanilla integrity verification framework where the prompt selection module Randomly picks and tests prompts from the library, using $n=50$ images generated from the model for each prompt. 
\textbf{Random\_v1} and \textbf{PromptLA\_v1} used the same threshold $\theta=0.25$, while \textbf{Random\_v2} and \textbf{PromptLA\_v2} used the same threshold $\theta=0.3$.

As shown in Table~\ref{tab:as-auc}, without using PromptLA algorithm, for integrity violations that are difficult to detect such as LoRA2, PM1, and SD-v1.4, the AUC decreased significantly, and the FPR increased substantially, meaning a high likelihood of classifying an intact model as being modified. 
The reason is that without PromptLA, it is hard to accurately estimate the distribution of generated images, so the verification is corrupted by the high bias.

As shown in Table~\ref{tab:as-cost}, for easier integrity violations, random strategies has a lower cost, as most prompts can provide sufficient discrimination. PromptLA has better efficiency for more challenging integrity violations that are harder to identify. 
Because PromptLA can quickly eliminate less effective prompts at a minimal cost, thus reaching a balance between accuracy and efficiency.

\textbf{PromptLA\_v1} and \textbf{PromptLA\_v2} are configurations selected based on extensive experimentation, which also turns out that a smaller $\alpha$
 resulted in a stricter hypothesis testing, a smaller FPR, yet more queries/runtime, while a lower $\theta$ 
 raised the FPR and smoothly increased the AUC. Therefore, the PromptLA can appropriately operate within a flexible range of configurations according to the defenders'  personalized balance between efficiency and accuracy.

\begin{table}[t]
    \centering
    \fontsize{8}{8.5}\selectfont
\begin{tabular}{cccccc}
\toprule
\multirow{2}{*}{\textbf{Method}}       & \textbf{FPR}            & \multicolumn{4}{c}{\textbf{Integrity Violation (AUC)}}                        \\ 
\cmidrule(lr){2-2}\cmidrule(lr){3-6}
             & SD-v1.5           & LoRA1          & LoRA2          & PM1          & v1.4           \\ 
             \midrule
Random\_v1   & 0.580          & 0.952          & 0.756          & 0.572          & 0.629          \\
\textbf{PromptLA\_v1} & \textbf{0.100} & \textbf{0.995} & \textbf{0.945} & \textbf{0.732} & \textbf{0.928} \\ 
\midrule
Random\_v2   & 0.350          & 0.982          & 0.886          & 0.654          & 0.768          \\
\textbf{PromptLA\_v2} & \textbf{0.250} & \textbf{1.000} & \textbf{0.995} & \textbf{0.690} & \textbf{0.953} \\ 
\bottomrule
\end{tabular}
    \caption{Comparison of AUC and FPR with and without using the PromptLA. The 'FPR' column is the false positive rate when we use the original model as a violation to verify.}
    \label{tab:as-auc}
\end{table}

\begin{table}[t]
    \centering
    \fontsize{8}{8.5}\selectfont
\begin{tabular}{cccccc}
\toprule
\multirow{2}{*}{\textbf{Method}} & & \multicolumn{4}{c}{\textbf{Integrity Violation (Cost)}}  \\ 
\cmidrule(lr){2-2}\cmidrule(lr){3-6}
             & \multicolumn{1}{c}{SD-v1.5}          & LoRA1          & LoRA2          & PM1         & v1.4           \\ 
             \midrule
Random\_v1   & \multicolumn{1}{c}{1890}          & \textbf{52} & \textbf{293} & \textbf{1159} & 1078           \\
\textbf{PromptLA\_v1} & \multicolumn{1}{c}{\textbf{1645}} & 134          & 551          & 1354          & \textbf{846} \\ 
\midrule
Random\_v2   & \multicolumn{1}{c}{2036}          & \textbf{52} & \textbf{375} & 1709          & 1695           \\
\textbf{PromptLA\_v2} & \multicolumn{1}{c}{\textbf{1410}} & 132         & 628          & \textbf{1375} & \textbf{859} \\ 
\bottomrule
\end{tabular}
    \caption{Comparison of Cost with and without using the PromptLA. The value of Cost is the average number of images generated per verification.}
    \label{tab:as-cost}
\end{table}

\subsection{Discussion}

The magnitude of integrity violations partially determines the difficulty of spotting them. 
However, different types of integrity violations can hardly be measured with respect to a unified metric regarding magnitude, either the distance in parameters or that in decision boundaries does not turn out to be conclusive. 
Due to a consistency concern, we studied this effect by confining the category of modifications to parameter-level modifications, where there is a well-defined order among all modification. 
As shown in Table~\ref{tab:extra}, when the degree of T2I diffusion model parameter modification gradually increased, so did the AUC of integrity verification. 
Meanwhile, the Cost decreased, which indicated that integrity violations of a larger magnitude are easier to detect. 
However, a theoretically unified and convincing metric on the magnitude of integrity violations to DNN, including T2I models, remains a challenge to be addressed.

\begin{table}[t]
    \centering
    \fontsize{8.5}{9}\selectfont
\begin{tabular}{cccccc}
\toprule
\textbf{Amplitude}    & \textbf{Euclidean} & \multicolumn{2}{c}{\textbf{PromptLA\_v1}} & \multicolumn{2}{c}{\textbf{PromptLA\_v2}} \\ 
\cmidrule(lr){3-4} \cmidrule(lr){5-6}
\textbf{Coefficient} & \textbf{Distance}  & AUC             & Cost           & AUC             & Cost           \\ 
\midrule
0.0010      & 110.19    & 0.732           & 1354           & 0.690           & 1375           \\
0.0015      & 165.24    & 0.865           & 1221           & 0.720           & 1465           \\
0.0020      & 220.31    & 0.933           & 533          & 0.920           & 746          \\
0.0025      & 275.37    & 0.950           & 303          & 0.975           & 349          \\
0.0030      & 330.48    & 0.993           & 148          & 0.997           & 148          \\
0.0035      & 385.51    & 1.000           & 126          & 1.000           & 127          \\ 
\bottomrule
\end{tabular}
    \caption{Integrity verification results vary when parameters are added with various levels of random noise. The noise was uniformly sampled from [0, 1) and scaled by a coefficient that modulates its amplitude. The amplitude coefficient ranged from 0.0010 to 0.0035, with the parameter differences between models measured using Euclidean distance.}
    \label{tab:extra}
\end{table}

\section{Conclusion}

In this paper, we propose the first integrity verification scheme of T2I diffusion models. 
We discern model modification via the KL divergence between the distributions of the features of generated images, and propose a prompt selection algorithm based on learning automaton (PromptLA). 
Extensive experiments demonstrate that our algorithm offers superior detection accuracy, efficiency, and generalization. 
PromptLA is also robust against image-level post-processing. 
The discussion in the experimental section regarding the impact of the degree of integrity tampering on detection results is currently limited to parameter modifications, yet it paves the way to more comprehensive metrics to measure the level of integrity violations. In our future work, we will explore optimizing prompts in continuous spaces and aim to extend the integrity verification framework to various complex generative tasks beyond text-to-image.

\bibliographystyle{ACM-Reference-Format}
\bibliography{sample-base}

\end{document}